\documentclass[conference,a4paper]{APSIPA2021}
\usepackage{amsmath}
\usepackage{graphicx}
\usepackage{multirow}
\usepackage{threeparttable}

\usepackage{geometry}
\usepackage{fancyhdr}
\usepackage{float}
\usepackage{amssymb}
\usepackage{hyperref}
\usepackage{bm}

\begin{document}

\title{PRTGaussian: Efficient Relighting Using 3D Gaussians with Precomputed Radiance Transfer}

\author{
\authorblockN{
Libo Zhang,
Yuxuan Han,
Wenbin Lin,
Jingwang Ling, and
Feng Xu
}
\authorblockA{
School of Software and BNRist, Tsinghua University\\
E-mail: zhanglb21@mails.tsinghua.edu.cn Tel: +86-15393595620
}
}


\maketitle
\pagestyle{empty}

\begin{abstract}
We present PRTGaussian, a realtime relightable novel-view synthesis method made possible by combining 3D Gaussians and Precomputed Radiance Transfer (PRT). 
By fitting relightable Gaussians to multi-view OLAT data, our method enables real-time, free-viewpoint relighting.
By estimating the radiance transfer based on high-order spherical harmonics, we achieve a balance between capturing detailed relighting effects and maintaining computational efficiency. 
We utilize a two-stage process: in the first stage, we reconstruct a coarse geometry of the object from multi-view images. In the second stage, we initialize 3D Gaussians with the obtained point cloud, then simultaneously refine the coarse geometry and learn the light transport for each Gaussian. 
Extensive experiments on synthetic datasets show that our approach can achieve fast and high-quality relighting for general objects.
Code and data are available at \href{https://github.com/zhanglbthu/PRTGaussian}{https://github.com/zhanglbthu/PRTGaussian}.
\end{abstract}

\section{Introduction}
Relightable view synthesis has played a crucial role in computer graphics and computer vision for a long time \cite{nrhints, relighable-3dgs, relightable-avatar, all-frequency}. 
It has many applications such as augmented reality and virtual object insertion. 
However, decoupling the lighting and reflectance information from the visual inputs and performing high-quality relighting is still slow, ill-posed and challenging.

Recently, some studies \cite{gir,relighable-3dgs,tensoir} try to employ inverse rendering techniques to explicitly estimate objects' intrinsic properties (\emph{i.e.} geometry and material) and scene lighting from images.
However, these methods struggle to model complex light transport, such as subsurface scattering and indirect illumination.
To tackle this problem, some methods \cite{neural-light-transport,nrhints,eyenerf} have attempted to directly model the light transport of objects using multi-view one-light-at-a-time (OLAT) datasets.
Nevertheless, such approaches often require cumbersome representation and intensive sampling, resulting in slow training and rendering speeds, severely limiting their application scenarios. Realtime relightable view-synthesis can improve interactive applications like VR and gaming, enabling dynamic, user-responsive lighting.

In this paper, we propose PRTGaussian, a novel framework for fast training and realtime relighting.
Given multi-view OLAT images of general objects, real-time view synthesis and relighting can be realized after efficient training.
Specifically, we use 3D Gaussians \cite{3dgs} with high-order spherical harmonics as the scene representations. 
Compared to implicit representations like NeRF \cite{nerf}, this explicit representation allows for more efficient forward rendering, resulting in faster training and inference speeds. 
The computation of object's appearance is based on the idea of precomputed radiance transfer. 
To be specific, for each Gaussian, we take its encoded position as input to regress its transport coefficients via a neural network, the final color is then obtained by combining this information with its albedo and lighting approximated by spherical harmonics.
However, jointly optimizing the geometry and the appearance of objects is challenging due to the high ambiguity and complexity involved in accurately capturing both aspects simultaneously. 
To mitigate this problem, we propose a two-stage training strategy. 
In the first stage, we construct a multi-view images of fixed illuminations from the OLAT dataset and obtain the position initialization of Gaussians from it using vanilla 3D Gaussian Splatting.
In the second stage, we initialize this point cloud as 3D Gaussians and proceed with further training, which refines the geometry of 3DGS and learns the light transport for each Gaussian.

In summary, the contributions of this paper can be outlined as follows:
\begin{itemize}
    \item We propose an efficient relighting framework that utilizes 3D Gaussians for scene representation and precomputed radiance transfer for appearance modeling.
    \item We introduce a two-stage training process, beginning with coarse point cloud reconstruction and 3DGS initialization from multi-view images, followed by the refinement of 3D Gaussians, along with the learning of light transport.
    \item We conduct extensive experiments to demonstrate that our method achieves high-quality relighting effects that are comparable with SOTA while maintaining fast rendering speeds.
\end{itemize}
\section{Related Work}
\begin{figure*}[tp]
    \centering
    \includegraphics[width=0.85\textwidth]{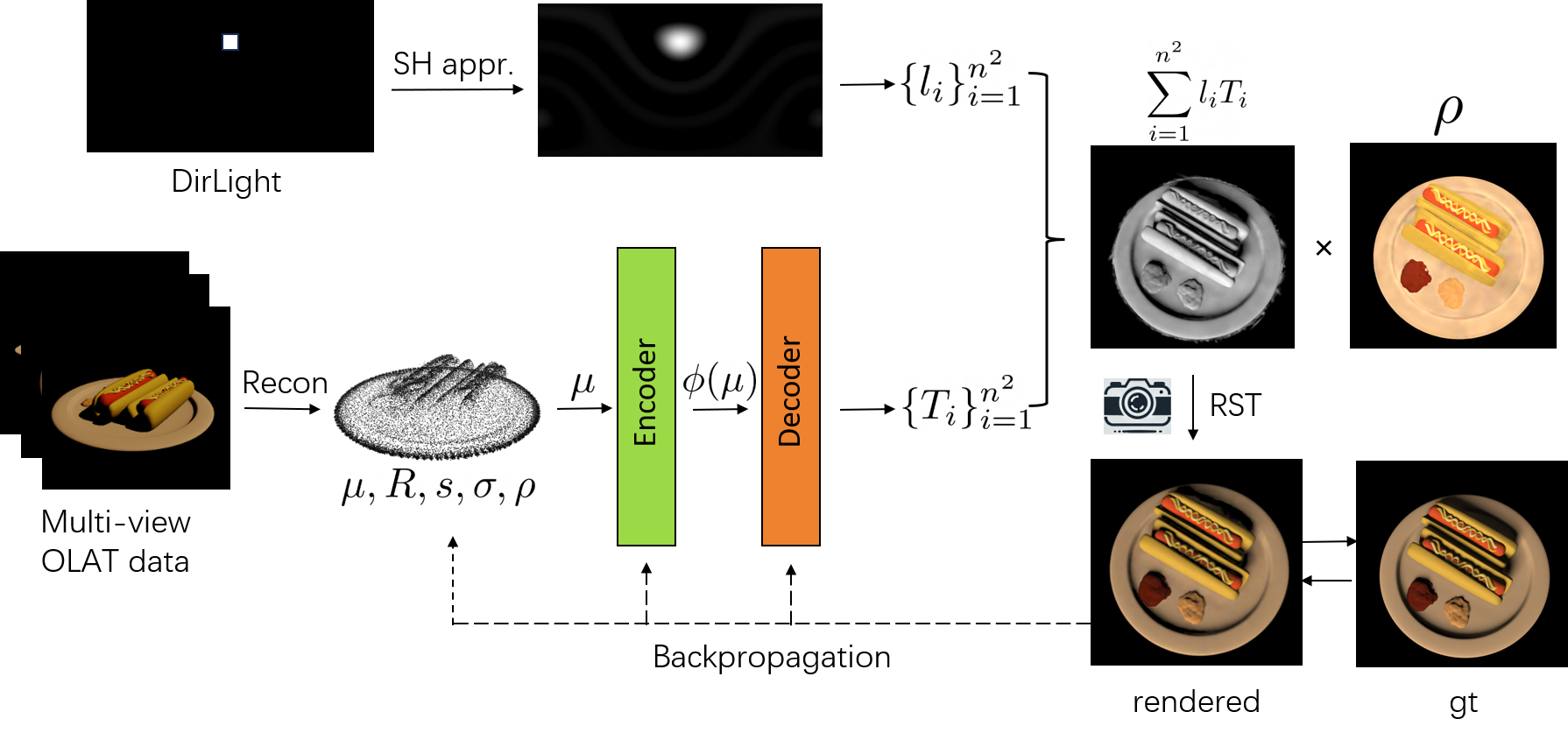}
    \caption{\textbf{Overview of Pipeline}. Given a multi-view OLAT dataset, our method first reconstruct the initial point cloud as the initialization for the 3D Gaussians. In the next stage, the color of each Gaussian can be obtained by combining the learned radiance transfer, albedo and the light source approximated by spherical harmonics (SH). We jointly optimize the attributes of the Gaussians as well as the encoder and decoder by minimizing the loss between rendered images and ground truth images.} 
    \label{fig:pipeline}
\end{figure*}

\subsection{3D Scene Representation}
Traditional scene representations \cite{lorensen1998marching, qi2017pointnet}, including mesh-based, voxel-based, and point cloud methods, have been widely used due to their simplicity and efficiency. 
However, these approaches often struggle with complex geometries and detailed appearances. 
To address these challenges, implicit representations \cite{nerf,neus} such as Neural Radiance Fields (NeRF) \cite{nerf} have emerged, providing continuous and high-resolution modeling capabilities that significantly enhance the quality and realism of 3D scene representations. 
Specifically, NeRF represents a continuous scene as a 5D vector-valued function, with the 3D spatial position and 2D viewing direction as inputs, and the corresponding color and volume density as outputs.
The image can then be rendered using volumetric rendering techniques. 
However, the sampling of rays and the use of a large Multi-Layer Perceptron significantly reduce its speed. 
Subsequent works \cite{mip-nerf,hu2022efficientnerf} have improved the quality and speed of NeRF, but their training and rendering time remain extremely high.

Recently, 3D Gaussians Splatting \cite{3dgs} has been introduced as an explicit scene representation. 
It uses 3D ellipsoids to represent scenes and employs a tile-based rasterization approach to render images, significantly improving training and rendering speeds while achieving high-quality novel view synthesis. 
The explicit nature of this representation also makes it more suitable for tasks such as scene editing and dynamic reconstruction.
In this work, we use 3D Gaussians (3DGS) to represent objects.
\subsection{Inverse Rendering}
The goal of inverse rendering is to decompose the scene's geometry, material, and lighting components from images. 
Under fixed lighting conditions, some methods \cite{gsir,relighable-3dgs} use differentiable rendering to estimate an object's depth, normal and material properties.
Since the forward process of these methods is often based on simple physical rendering models such as the Cook-Torrance microfacet model \cite{cook}, they can handle only limited geometries and materials, such as subsurface scattering, transparency, and anisotropic materials.

Other methods perform inverse rendering based on known multi-illumination conditions. 
NRHints \cite{nrhints} uses unstructured multi-view data with moving point light sources, employing two MLPs to separately represent the object's geometry and appearance.
Alternative approaches \cite{neural-light-transport, codec_avatar} learn appearance representations based on one-light-at-a-time (OLAT) data, but they typically focus on specific objects, such as human bodies and faces. 
We also utilize the OLAT dataset, but our method achieve lighting decoupling for general objects without relying on any priors.

\subsection{Precomputed Radiance Transfer}
In traditional computer graphics, rendering global illumination effects for a scene is costly due to the extensive ray tracing and numerous light bounce computations required. 
To achieve real-time rendering with global illumination, Sloan et al. \cite{sloan2002precomputed} introduced the method of precomputed radiance transfer (PRT), which precomputes the coefficients of incident light projected onto spherical harmonics bases and the transfer vector at each point on the surface. 
The transfer vector, also represented by spherical harmonics coefficients, describes how each point on the surface interacts with incident light, including global transport effects such as reflections and shadows. 
At runtime, the light vector and transfer vector are integrated to obtain the radiance at each point.
To address the low-frequency limitations of spherical harmonics, \cite{nonlinear}\cite{all-frequency} introduced new representations such as nonlinear Gaussian functions and spherical Gaussians based on this foundation. 
However, these methods rely on known geometry and material properties.

We aim to model the radiance transfer at each point solely from images. 
Specifically, we use two networks to separately encode and decode the radiance transfer vector.

\section{Method}
\begin{figure*}[t]
    \centering
    \includegraphics[width=1.0\textwidth]{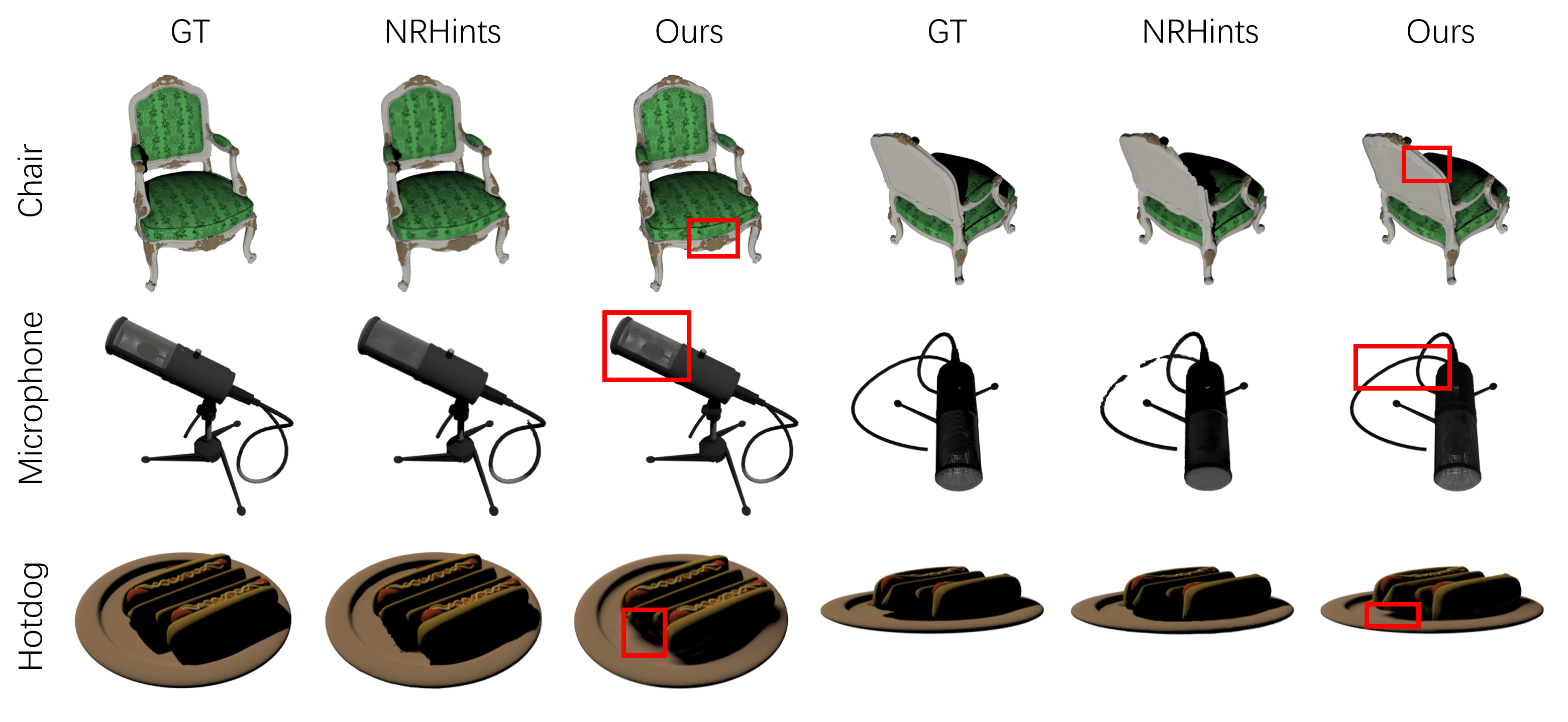}
    \caption{\textbf{Qualitative Comparisons.} We performed qualitative comparisons with the state-of-the-art method NRHints\cite{nrhints}. Here we present the ground truth and the rendering results of different methods for three subjects under two novel lighting and view conditions.}
    \label{fig:quantatitive}
\end{figure*}
\begin{figure}[h]
    \centering
    \includegraphics[width=0.65\linewidth]{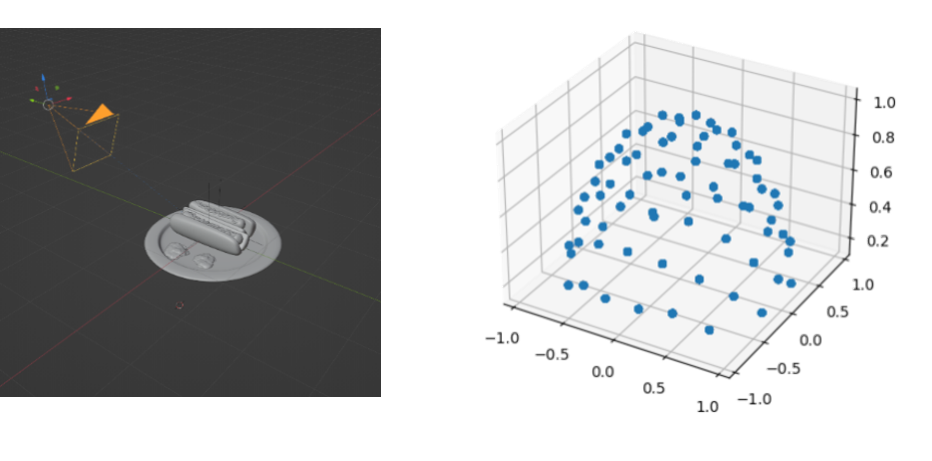}
    \caption{Data Synthesis Setup. \textit{Left} represents the setup in Blender and \textit{Right} represents the sampled positions of the cameras and light sources.}
    \label{fig:data_acquisition}
\end{figure}

Given the multi-view OLAT data, our goal is to achieve free-viewpoint relighting of the object. 
An overview of our method is shown in Fig.\ref{fig:pipeline}. 
In the first stage, we reconstruct the initial positions for the 3D Gaussians. In the second stage, we refine the properties of 3DGS and learn the radiance transfer.
In this section, we provide details on data acquisition (Sec.\ref{sec:data_acquisition}), initial geometry reconstruction (Sec.\ref{sec:igr}) and radiance transfer learning (Sec.\ref{sec:rtl}).

\subsection{Data Acquisition}
We use a one-light-at-a-time (OLAT) dataset similar to the light-stage \cite{lightstage} setup, where camera and light information are known. 
We synthesize a multi-view OLAT dataset for general objects in Blender. 
Specifically, we uniformly sample 25 camera positions and 200 lighting positions on the upper hemisphere of the object. 
The lighting is considered as directional light, with the detailed setup shown in Fig.\ref{fig:data_acquisition}.
\label{sec:data_acquisition}

\subsection{Initial Geometry Reconstruction}
\label{sec:igr}
We use a set of 3D Gaussians (3DGS) to represent the object's geometry. Each Gaussian can be defined as 
\begin{align}
    g_k = \{\bm{\mu}, \mathbf{R}, \bm{s}, \sigma, \bm{\rho} \}
\end{align}
where $\bm{\mu}\in \mathbb{R}^3$ is the 3D position, $\mathbf{R}\in SO(3)$ is the rotation matrix, $\bm{s}\in \mathbb{R}^3$ is the per-axis scale factors, $\sigma \in \mathbb{R}$ is the opacity value and $\bm{\rho} \in \mathbb{R}^3$ is the albedo. The covariance matrix of each Gaussian can be represented as:
\begin{align}
    \Sigma = \mathbf{R}\text{diag}(\bm{s})\text{diag}(\bm{s})^T\mathbf{R}^T
\end{align}
In the splatting process, 3D Gaussians are first projected onto the 2D plane,
\begin{align}
    \Sigma' = JV\Sigma V^TJ^T
\end{align}
where $J\in \mathbb{R}^{2\times3}$ is the Jacobian of the projective transformation, $V\in \mathbb{R}^{3\times3}$ is the viewing transformation, and $\Sigma' \in R^{2\times2}$ is the covariance matrix of the projected 2D Gaussians. 
The color $\mathbf{C}_p$ of each pixel can be computed using accumulated volumetric rendering as follows:
\begin{align}
    \mathbf{C}_p = \sum_{k\in N} \bm{c_k}\alpha_k \prod_{j=1}^{k-1}(1-\alpha_j)
\end{align}
where $k$ is the index of N ordered Gaussians, $\alpha_k \in \mathbb{R}$ is estimated from the 2D covariance matrix and the opacity of each Gaussian, and $c_k \in \mathbb{R}^3$ is the color, which is determined by the albedo, radiance transfer, and incident light.

Due to the high ambiguity in jointly optimizing geometry and appearance, we first reconstruct a coarse geometry of the object from multi-view images under uniform lighting. 
The uniformly illuminated multi-view data is obtained by summing and averaging the illumination for each viewpoint.
Then, We use the vanilla 3D Gaussian Splatting\cite{3dgs} method for coarse geometry reconstruction from these images. 
The generated point cloud is applied for Gaussian initialization in the next stage.
\subsection{Radiance Transfer Learning}
\begin{figure*}[tp]
    \centering
    \includegraphics[width=1.0\textwidth]{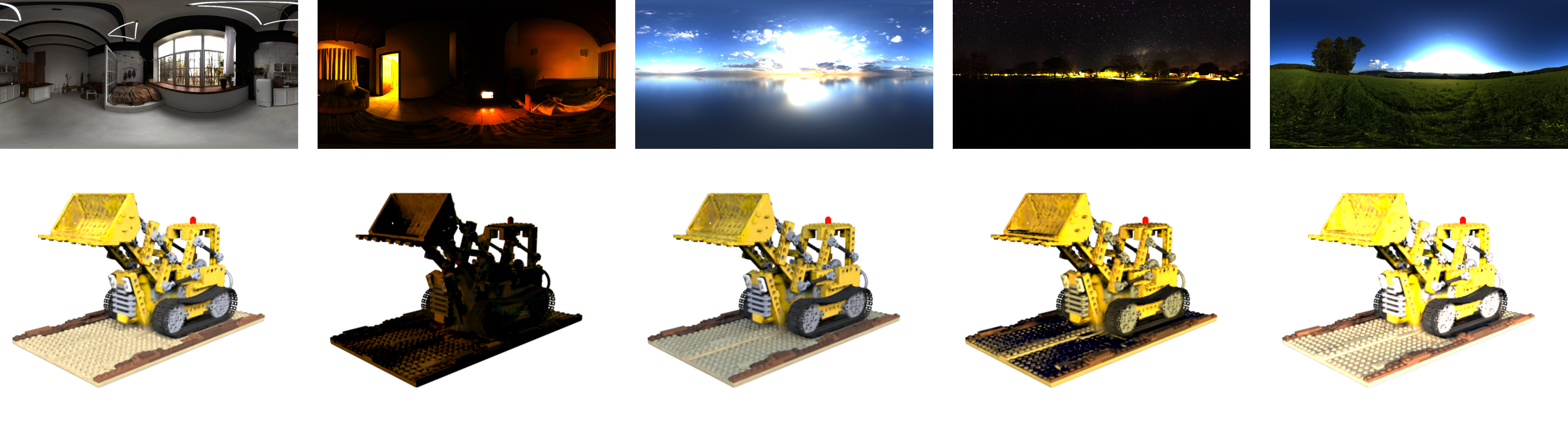}
    \caption{\textbf{Qualitative Results.} The top shows different environment maps, while the bottom displays the corresponding results using our model trained on the OLAT dataset.}
    \label{fig:qualitative}
\end{figure*}
\label{sec:rtl}
In the second stage, we learn the radiance transfer for each 3D Gaussian. The basic rendering equation can be written as:
\begin{align}
    L(o)=\int_{\Omega}L(i)V(i)\rho(i,o)\max(0,\text{n}\cdot i)di
    \label{equ: rendering equation}
\end{align}
where $i$ and $o$ represent the incident and outgoing directions, while $L,V,\rho,\text{n}$ represent the lighting, visibility, albedo, and normal properties, respectively. 

In this work, we primarily focus on diffuse materials where each Gaussian has a specific albedo $\rho(i,o)=\rho$. Based on the concept of Precomputed Radiance Transfer (PRT), the light and radiance transfer can be precomputed as follows:
\begin{align}
    &L(i) = \sum_{j=1}^{n^2}l_j B_j(i)\\
    &T_j = \int_{\Omega}B_j(i)V(i)max(0, \text{n}\cdot i)di
\end{align}
where $n$ is the order of the spherical harmonics, $B_j$ is the spherical harmonics basis, $l_j\in \mathbb{R}^{n^2}$ is the coefficient of the light source projected onto SH space, and $T_j \in \mathbb{R}^{n^2}$ is the radiance transfer term. Therefore, equation \ref{equ: rendering equation} can be rewritten as:
\begin{align}
    L(o) = \rho \sum_{j=1}^{n^2}l_j T_j
\end{align}

The light coefficient $l_j$ can be obtained from the input lighting by direct projection.
The radiance transfer term $T_i$ is encoded with a neural network.
Specifically, we first encode the position $\mu$ of each Gaussian into a higher-dimensional vector \cite{instantngp} $\phi(\mu)$, and then decode it using an MLP to obtain the corresponding $T_i$.

During the training process, the rendered image is compared with the ground truth image to calculate the loss:
\begin{align}
    L = (1-\lambda) L_1 + \lambda L_{D-SSIM}
\end{align}
we use the gradient of this loss to update the parameters of the encoder and decoder, while simultaneously refining all the properties of each Gaussian.

\section{Experiments}
\begin{figure*}[htp]
    \centering
    \includegraphics[width=1.0\textwidth]{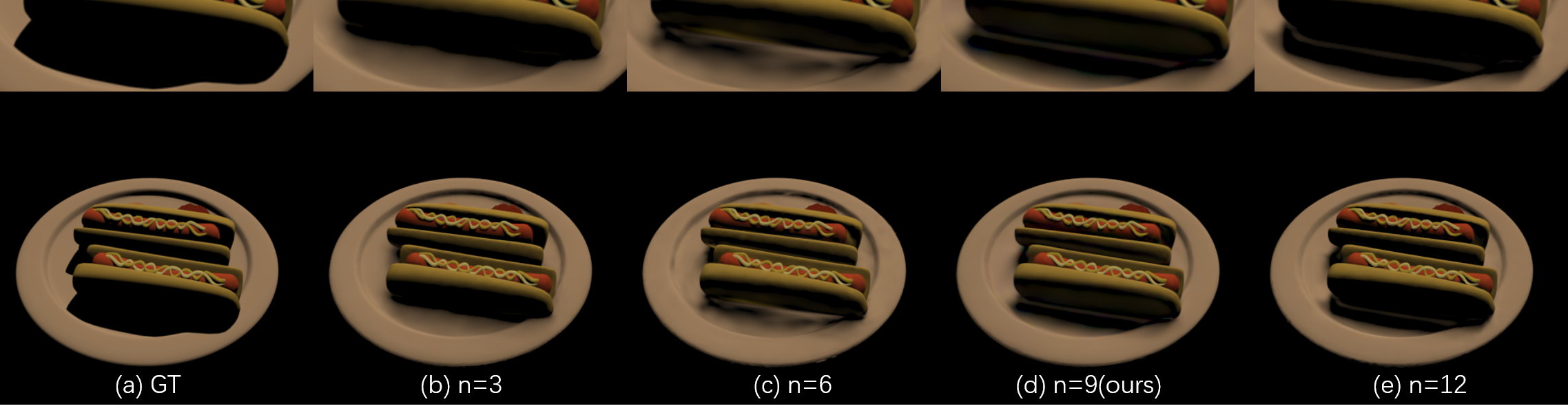}
    \caption{\textbf{Ablation Study: The order of the SH basis.} Compared to the ground truth (a), using higher-order SH coefficients can enhance the sharpness of shadows. To balance quality and efficiency, we choose $n=9$ in our method.}
    \label{fig:sh_order}
\end{figure*}
\subsection{Implementation Details}
We implemented our method in Python using the PyTorch framework. 
Our encoder uses hash-grid encoding \cite{instantngp} with the number of levels, number of features per level, and hash table size set to 32, 8, and $2^{20}$, respectively. 
Our decoder employs a four-layer fully connected MLP, with each layer consisting of 256 neurons. 
To capture higher frequency information, we set the order of the spherical harmonics basis to $n=9$. 
During training, we used the Adam optimizer and set $\lambda$ to $0.2$. 
The entire training process was conducted on an NVIDIA RTX 4090, with a total training time of approximately 15 minutes per scene.
\subsection{Quantitative Results}
We conducted a comprehensive comparison against the state-of-the-art field-based method NRHints \cite{nrhints}, which uses volume rendering and represents geometry in SDF. 
Tab. \ref{tab:quantatitive_comparison} summarizes the quantitative comparison results on synthetic data, with each metric averaged over 3 subjects.
We also compared the rendering speed (per frame) and training time of these two methods.
All comparison experiments are conducted under the same conditions and on the same equipment. 
\begin{table}[h]
\centering
\caption{\textbf{Quality Comparison.}}
\label{tab:quantatitive_comparison}
\begin{tabular}{l|ccc|cc}
\hline
 & \textbf{PSNR $\uparrow$} & \textbf{SSIM $\uparrow$} & \textbf{LPIPS $\downarrow$} & \textbf{render} & \textbf{train}\\
\hline
NRHints \cite{nrhints} & 30.77 & \textbf{0.9729} & 0.0263 & 17.95s & 28h\\
Ours       & \textbf{33.63} & 0.9438 & \textbf{0.0255} & \textbf{0.0030s} & \textbf{15m}\\
\hline
\end{tabular}
\end{table}

The 3DGS-based scene representation enables our method to facilitate efficient training and inference while simultaneously capturing fine details of apperances.
The experimental results demonstrate that our approach significantly improves training and rendering speeds compared to previous method, while achieving comparable rendering quality.

\subsection{Qualitative Results}
We also conducted a qualitative comparison with the current top-performing technique (NRHints) as shown in Fig. \ref{fig:qualitative}. Thanks to the scene representation capability of 3DGS and the learning of radiance transfer, our method can recover fine details in both geometry and appearance, such as \textit{Chair's patterns, Microphone's wires and Hotdog's shadows}, which other methods fails to reproduce.

Once trained on the multi-view OLAT dataset, our method can achieve relighting under arbitrary lighting conditions. In Fig. \ref{fig:qualitative}, we present the qualitative results of HDRI relighting. 
For environment maps, our approach first projects them into SH space and then performs efficient and high-quality relighting.

\subsection{Ablation Studies}

\begin{figure}[h]
    \centering
    \includegraphics[width=1.0\linewidth]{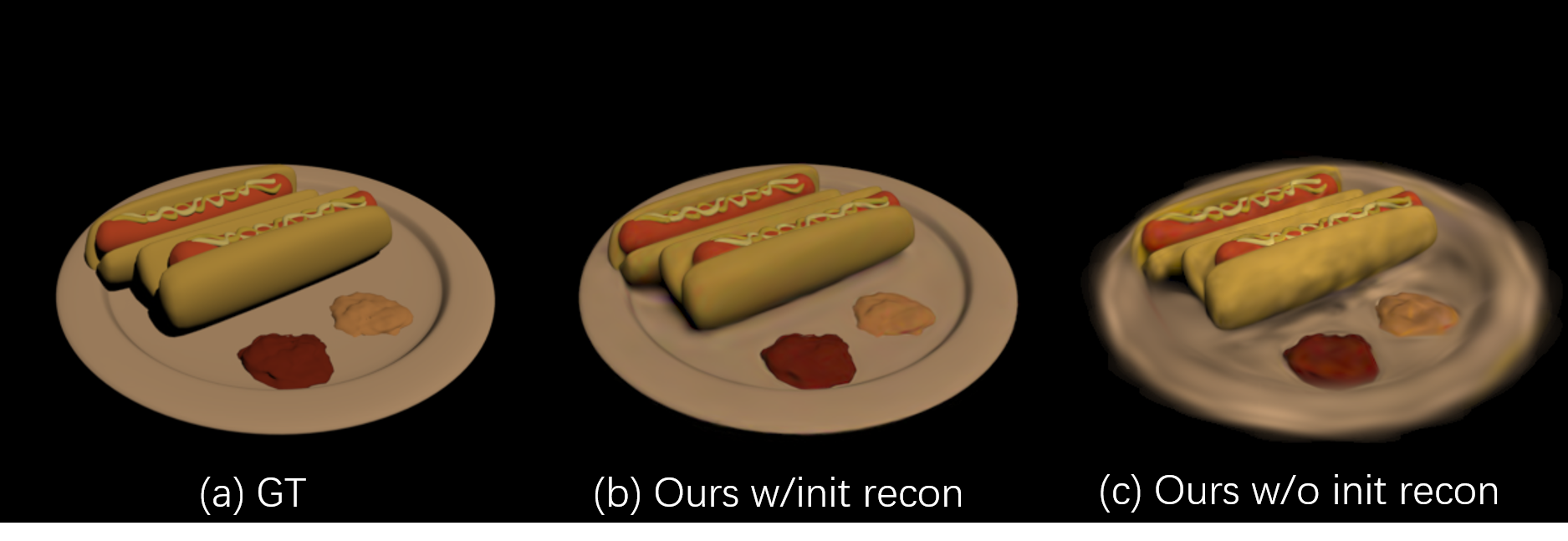}
    \caption{\textbf{Ablation Study: Initial Geometry Reconstruction.} Compared to the ground truth (a), using initial geometry reconstruction (b) leads to more accurate rendering results than randomly initialized geometry (c).}
    \label{fig:init_geo}
\end{figure}
\begin{figure}[h]
    \centering
    \includegraphics[width=1.0\linewidth]{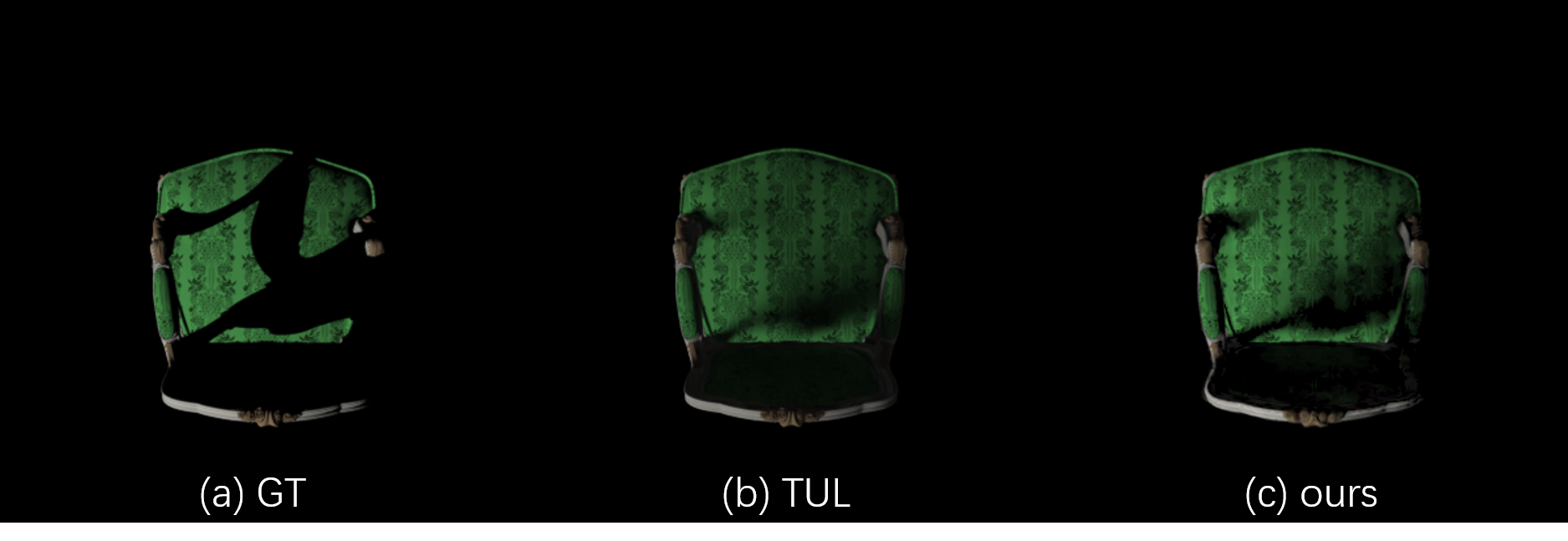}
    \caption{\textbf{Limitations.} The left (a) is the ground truth, the middle (b) is the results rendered in Blender using directional light approximated by SH, which represents our theoretical upper limit (TUL), and the right (c) shows the results obtained by training our model with the ground truth. It can be seen that our method is inherently difficult to achieve high-frequency effects such as hard shadows limited by the low-frequency nature of SH basis.}
    \label{fig:limitation}
\end{figure}

In this section, we provide several ablation studies to validate the key design choices of our method. 

\textbf{Initial Geometry Reconstruction.} 
In stage 1, we performed initial point cloud reconstruction to provide a better initialization for 3D Gaussians (3DGS). 
We compare our approach with one that does not use initial reconstruction but instead initializes 3DGS by sampling them uniformly in a cube to verify the necessity of this design. 
Fig. \ref{fig:init_geo} shows our method achieves significantly better results. 
This improvement is due to the high ambiguity in inverse rendering and the challenge of simultaneously optimizing the geometry and appearance of 3DGS using OLAT datasets.

\textbf{High Order SH Basis.} 
In our method, both the lighting and radiance transfer are represented using spherical harmonics (SH), and the number of SH basis coefficients is positively correlated with the square of its order. 
To capture high-frequency effects (such as shadows) while controlling the number of SH coefficients to improve computational efficiency and fit GPU memory, our method uses spherical harmonics up to the 8th order (nine orders in total). 

We compared the results with those of different orders (n=3, 6, 12) and found that our method balances the efficiency and quality. 
Fig.\ref{fig:sh_order} shows that when the order is too low, it is difficult to represent high-frequency information such as shadows, and when the order is too high, the memory and computational requirements increase significantly with minimal improvement in rendering quality.

\section{Limitations}

Although we have accelerated the training and inference speeds of existing relighting methods and achieved decent rendering results, our approach still has some limitations and shortcomings.

One major drawback is that we currently only consider diffuse materials and do not consider specular highlights and other view-dependent effects. 
Incorporating the view direction as a condition in the pipeline holds promise for addressing this issue in the future.

Another drawback is that although we use high-order SH to approximate the information of light and radiance transfer, our method still struggles to reproduce high-frequency rendering effects such as the sharp edges in hard shadows (as shown in Fig.\ref{fig:limitation}). 
This is due to the inherently low-frequency nature of spherical harmonics. 
Explicitly considering the visibility of 3DGS in future work will help alleviate this situation.
\section{Conclusions}
In this study, we present PRTGaussian, a novel relighting framework that combines 3D Gaussian Splatting and Precomputed Radiance Transfer to achieve efficient and high-quality relighting. 
Our approach demonstrates significant improvements in training and rendering speeds compared to existing techniques and efficiently utilizes multi-view OLAT data for realistic relighting.
While our method excels in these areas, it is currently limited to diffuse materials and has difficulty capturing high-frequency details. Future work could address these limitations by integrating view-dependent effects and enhancing the representation of high-frequency components.
\bibliographystyle{IEEEtran} 
\bibliography{main}

\end{document}